\documentclass{article}
\usepackage{iclr2027_conference,times}
\usepackage{amsmath,amssymb}
\usepackage{booktabs}
\usepackage{graphicx}
\usepackage[table]{xcolor}
\usepackage{tikz}
\usepackage{hyperref}
\usepackage{url}
\usepackage{xurl}
\usepackage{multirow}
\definecolor{GtwoOPDFill}{HTML}{DCECF8}
\definecolor{GtwoOPDInk}{HTML}{145D8D}
\definecolor{GtwoRLFill}{HTML}{FCE6D2}
\definecolor{GtwoRLInk}{HTML}{8C481D}
\definecolor{GtwoMissingInk}{HTML}{777777}

\newcommand{\GtwoRL}{\textcolor{GtwoRLInk}{RL}}
\newcommand{\GtwoOPD}{\textcolor{GtwoOPDInk}{OPD}}
\usepackage{threeparttable}
\usepackage{wrapfig}
\usepackage{tabularx}
\usepackage{array}
\usepackage{booktabs}
\usepackage{needspace} % preamble
\usepackage{booktabs}
\usepackage{multirow}
\usepackage{xcolor}

\definecolor{TblOPDInk}{HTML}{145D8D}
\definecolor{TblRLInk}{HTML}{8C481D}
\newcommand{\TblOPD}{\textcolor{TblOPDInk}{OPD}}
\newcommand{\TblRL}{\textcolor{TblRLInk}{RL}}
\newcommand{\TblOPDWin}[1]{\textcolor{TblOPDInk}{\textbf{#1}}}
\newcommand{\TblRLWin}[1]{\textcolor{TblRLInk}{\textbf{#1}}}

\newcommand{\TblBody}{\normalsize\renewcommand{\arraystretch}{1.0}}

\hypersetup{
  pdftitle={No Task Vector Is an Island: A Comprehensive Study on the Composability of Task Vectors from On-Policy Distillation},
  pdfauthor={Jingang Zhou, Feiyu Han, Han Zhu, Yuyi Zhou, Ruiyang Zhang, Jian Xu, Sirui Gao, Qingpei Guo, Xu-Yao Zhang}
}

\newcommand{\aff}[1]{\textsuperscript{\normalfont #1}}

\title{%
  \begin{minipage}{\textwidth}
    \centering
    \normalfont
    \makebox[\linewidth][c]{%
      \raisebox{-0.20\height}{%
        \includegraphics[height=0.65cm]{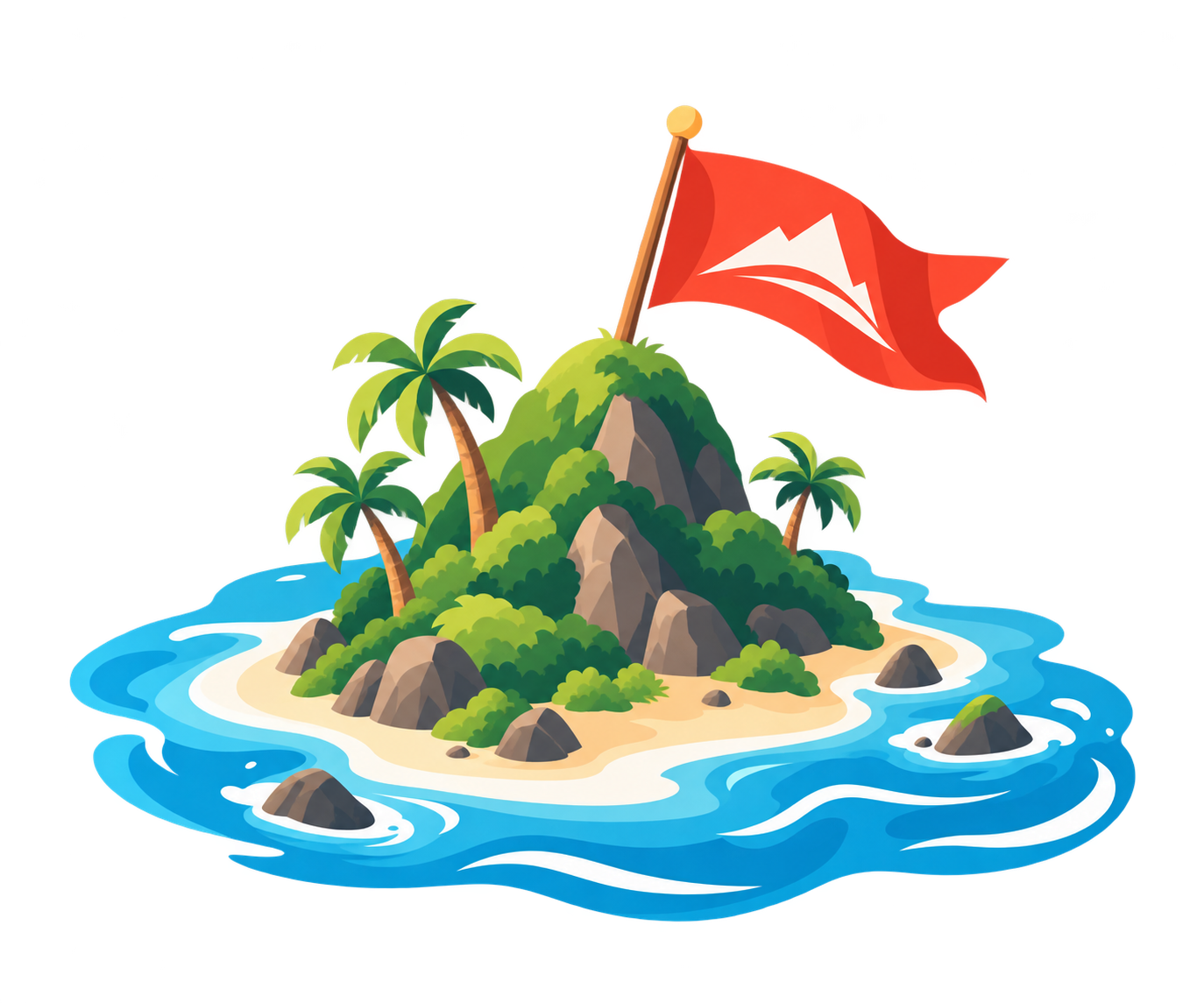}%
      }%
      \hspace{0.55em}%
      {\fontsize{17.28}{20}\selectfont\scshape
       No Task Vector Is an Island:}%
    }\\[5pt]
    {\fontsize{13}{16}\selectfont\scshape
      A Comprehensive Study on the Composability\\
      of Task Vectors from On-Policy Distillation\endgraf
    }
  \end{minipage}%
}

\author{%
  \parbox[t]{\dimexpr\textwidth-2\tabcolsep\relax}{%
    \centering
    \normalfont\fontsize{10}{13}\selectfont
    {\bfseries
      Jingang Zhou\aff{1,2}\quad
      Feiyu Han\aff{2}\quad
      Han Zhu\aff{1,2}\quad
      Yuyi Zhou\aff{1,2}\quad
      Ruiyang Zhang\aff{3}\\[2pt]
      Jian Xu\aff{1,2}\quad
      Sirui Gao\aff{3}\quad
      Qingpei Guo\aff{3}\quad
      Xu-Yao Zhang\aff{1,2,*}\endgraf
    }
    \vspace{5pt}
    {\normalfont\fontsize{9}{11.5}\selectfont
      \aff{1}\,Institute of Automation, Chinese Academy of Sciences\\
      \aff{2}\,University of Chinese Academy of Sciences\quad \qquad\aff{3}\,Ant Group\endgraf
    }
    \vspace{3pt}
    {\normalfont\fontsize{9}{11}\selectfont
      \aff{*}\,Corresponding author\endgraf
    }
  }%
}

\newcommand{\norm}[1]{\left\lVert #1\right\rVert}

\newif\ifdraftspaces
\draftspacestrue

\definecolor{findingbackground}{HTML}{EAF6ED}
\definecolor{findingborder}{HTML}{9FC6AB}
\definecolor{takeawaybackground}{HTML}{FBE8ED}
\definecolor{takeawayborder}{HTML}{E8A8BA}
\definecolor{TblOPDInk}{HTML}{145D8D}
\definecolor{TblRLInk}{HTML}{8C481D}
\definecolor{DeltaPos}{HTML}{3F7D4A}
\definecolor{DeltaNeg}{HTML}{A14A4A}
\definecolor{DeltaZero}{HTML}{777777}

\newcommand{\DeltaPosText}[1]{{\normalsize\textcolor{DeltaPos}{(#1)}}}
\newcommand{\DeltaNegText}[1]{{\normalsize\textcolor{DeltaNeg}{(#1)}}}

\newcommand{\ScoreWithPosDelta}[3][]{#2#1\,\DeltaPosText{+#3}}
\newcommand{\ScoreWithNegDelta}[3][]{#2#1\,\DeltaNegText{-#3}}

\newcounter{finding}
\newcommand{\finding}[2]{%
  \par\addvspace{0.5\baselineskip}%
  \begingroup
  \noindent\begin{tikzpicture}[baseline=(findingbox.base)]
    \node[draw=findingborder,fill=findingbackground,
      line width=0.4pt,rounded corners=4pt,inner sep=6pt,outer sep=0pt,
      text width=\dimexpr\linewidth-12.8pt\relax,align=justify]
      (findingbox) {%
      \refstepcounter{finding}\label{#1}%
      \textbf{Finding~\thefinding.}\ #2%
    };
  \end{tikzpicture}%
  \par\endgroup
  \addvspace{0.5\baselineskip}%
}

\newcommand{\takeaway}[1]{%
  \par\addvspace{0.5\baselineskip}%
  \begingroup
  \noindent\begin{tikzpicture}[baseline=(takeawaybox.base)]
    \node[draw=takeawayborder,fill=takeawaybackground,
      line width=0.4pt,rounded corners=4pt,inner sep=6pt,outer sep=0pt,
      text width=\dimexpr\linewidth-12.8pt\relax,align=justify]
      (takeawaybox) {\textbf{Takeaway.}\ #1};
  \end{tikzpicture}%
  \par\endgroup
  \addvspace{0.5\baselineskip}%
}

\newcommand{\paperfigure}[2][]{%
  \IfFileExists{#2}{\includegraphics[#1]{#2}}{%
    \fbox{\parbox[c][28mm][c]{0.92\linewidth}{%
      \centering\small\textbf{Figure file not supplied}\\[4pt]
      \texttt{\detokenize{#2}}}}%
  }%
}
\iclrfinalcopy
\begin{document}
\maketitle
\lhead{Preprint}

\begin{abstract}
Task vectors provide a simple mechanism for composing learned capabilities
through model merging. However, the composability of task vectors produced
by on-policy distillation (OPD) remains largely unexplored. OPD trains a
student using teacher feedback on student-generated trajectories, yielding
parameter updates that differ from those produced by the teacher model, usually by reinforcement learning
(RL). \textit{We therefore ask whether OPD task vectors can complement their
RL teacher updates and compose effectively across tasks.} Across five domains and two model architectures, we find evidence for both
forms of composability. \textbf{Within a task}, merging OPD and RL task
vectors can outperform both constituent models, even when the OPD student
is weaker than its RL teacher. \textbf{Across tasks}, OPD task-vector
compositions achieve higher average scores than corresponding RL
compositions in seven of eight backbone--merging-rule comparisons.
Parameter-space analyses reveal substantial non-collinearity between
OPD and RL updates. Experiment in \textsc{Code} domain on \textsc{SmolLM3-3B} shows that the combined direction outperforms either
constituent direction at the tested global update norm, supporting
directional complementarity in this configuration. Across tasks, OPD
updates also show lower overlap among the top-$10\%$ feed-forward
channels ranked by update energy. Together, these results show that weaker standalone performance does not
imply weaker task-vector composability. OPD task vectors can complement
stronger RL teacher updates and combine effectively across tasks,
highlighting composability as a distinct property for understanding and
evaluating post-training updates.
\end{abstract}

\begin{figure}[!h]
\centering
\includegraphics[width=\linewidth]{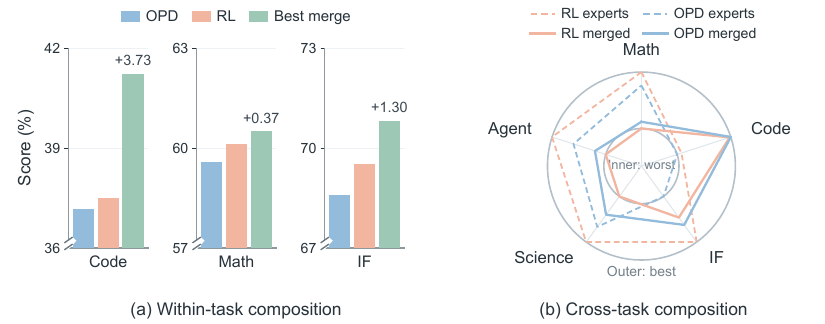}
\caption{\textbf{OPD task-vector composition on \textsc{Qwen3-4B}.}
(a) Best within-task OPD--RL merges; labels show gains over RL (pp).
(b) Specialists (dashed) and TIES merges (solid); radial scores are
min--max normalized separately for each task.}
\label{fig:intro-overview}
\end{figure}

% BEGIN SOURCE: sections/introduction.tex
\section{Introduction}
\label{sec:intro}

Post-training develops general-purpose language models into domain
experts. Training with human feedback improves instruction following
\citep{ouyang2022instructgpt}, while reinforcement learning with
verifiable rewards (RLVR) strengthens mathematical reasoning and code
generation \citep{deepseekai2025r1}.
Consolidating experts trained from a shared base is therefore a
practical challenge. Parameter-space merging combines their
\emph{task vectors}---the displacements from the common base---through
addition, interpolation, or other merging rules, without further
gradient-based training \citep{ilharco2023taskarithmetic}.
Multi-teacher on-policy distillation (OPD) instead transfers teacher
behavior into a shared student through dense supervision on
student-generated trajectories
\citep{agarwal2024gkd,ma2026mopd,gao2026openmopd}.
The compatibility of task vectors depends on how these updates are
learned: RL-trained experts can exhibit different conflict patterns
from supervised fine-tuning (SFT) experts \citep{ren2026feast}, and
CAMFT explicitly encourages updates to occupy coordinates with lower
cross-task conflict during fine-tuning \citep{zhou2026camft}.
These findings motivate studying how the training procedure shapes
mergeability alongside the choice of merging algorithm.

OPD provides a useful setting for studying how post-training shapes
mergeability: it trains on student-generated trajectories using
token-level feedback from a stronger teacher \citep{agarwal2024gkd, zhou2026acopd}.
The construction of this supervision matters. The parameter-space analyses
suggest that OPD produces update structures distinct from standard RL
or SFT \citep{shen2026opdgeometry,yu2026densesupervision}.
An OPD student's task vector may therefore differ from its RL teacher's.
We ask: \emph{how composable are the task vectors produced by OPD?}
Figure~\ref{fig:intro-overview} illustrates the two relationships we examine.
\textbf{Within the same task}, we test whether merging a weaker OPD
student with its RL teacher can outperform both constituent models,
which would reveal complementarity beyond selecting the stronger
endpoint. \textbf{Across different tasks}, we compare independently
trained OPD and RL task vectors to test whether OPD reduces interference
when combining domain capabilities.

We investigate these questions across five domains and two model
architectures.
\textsc{SmolLM3} covers \textsc{mathematical reasoning},\textsc{ code generation}, and \textsc{instruction
following}, while \textsc{Qwen} additionally includes \textsc{Science} and \textsc{Agent} tasks.
Across these settings, we observe two main empirical phenomena.
First, \textbf{within a task, merging an OPD task vector with its own
RL teacher's task vector can outperform both constituent models}.
On \textsc{Qwen3-4B} in \textsc{Code}, the OPD student and RL teacher score $37.18\%$
and $37.51\%$ (Table~\ref{tab:opd-cross-task-results}); their
within-task TIES composition reaches $41.24\%$
(Table~\ref{tab:opd-within-task}), improving on the teacher by $3.73$ pp.
Thus, a weaker OPD student can contribute useful parameter changes
to its teacher. The results span multiple tasks and merging operators
(Section~\ref{sec:opd-within-task}).

Second, \textbf{across tasks, OPD task vectors generally compose more effectively
than corresponding RL task vectors}.
When multiple domain-specific updates are merged from the same base
model, OPD-based compositions achieve higher overall scores in seven of
the eight backbone--merging-rule comparisons, with \textsc{SmolLM3} Raw Sum as
the exception. Across all eight comparisons, OPD has a more favorable
change from its own standalone specialist average.
This distinction is useful because final merged performance alone can
hide asymmetric task degradation: two merging methods may obtain
similar averages while preserving their constituent experts very
differently.
Our results therefore evaluate both the final multi-task model and the
performance drop relative to each method's own single-task specialists.
Across the five-task, this comparison consistently favors
the compositional behavior of OPD updates, although the magnitude of
the advantage varies across domains.

We further connect these performance differences to the geometry of the
learned updates.
Within the same task, OPD updates have substantial components orthogonal
to the RL direction, although non-collinearity alone does not establish
a functional benefit. On \textsc{SmolLM3} in \textsc{Code}, the combined update outperforms both
constituent directions when each is rescaled to the same global norm
(Section~\ref{sec:within-task-complementarity}), supporting directional
complementarity at the tested norm.
Across tasks, we find a complementary pattern.
Direction retention improves for \textsc{Code}, \textsc{Science}, and \textsc{Agent}, but declines
for \textsc{Math} and \textsc{IF}. These diagnostics describe task-dependent interactions;
they do not establish a general relationship between alignment and
performance retention.
Together, these findings suggest that OPD does more than transfer
teacher performance into a student: it can reshape the resulting task
vector in ways that make learned capabilities easier to compose.

\begin{samepage}
Our contributions are:
\begin{itemize}
    \item We systematically study the \textbf{composability of OPD task
    vectors}, covering both within-task OPD--RL composition and
    cross-task multi-expert merging across five domains and two model
    architectures.

    \item To our knowledge, we are the first to show two complementary empirical properties of OPD updates: an OPD task vector can \textbf{strengthen the RL expert from which it was distilled}, and multiple OPD task vectors can \textbf{retain their specialist capabilities more effectively} than corresponding RL task vectors under cross-task merging.

    \item We characterize OPD update geometry and test directional complementarity with norm-matched controls, alongside task-dependent patterns of overlap and direction retention across tasks.
\end{itemize}
\end{samepage}
\section{Preliminaries}
\label{sec:preliminaries}

\subsection{Post-Training}

We focus on reinforcement learning with verifiable rewards (RLVR) and
policy-gradient on-policy distillation (OPD). RLVR derives its learning
signal from response-level answer checks or executable tests
\citep{deepseekai2025r1,shao2024deepseekmath}.
In OPD, a fixed teacher provides token-level feedback on student-generated
trajectories. We follow \citet{li2026rethinking}, using either sampled-token
feedback or supervision over multiple top-$k$ candidates.
The two methods share a closely related policy-gradient structure:
RLVR optimizes verifier-defined outcomes, while OPD aligns the student
with its teacher through token-level feedback.
Appendix~\ref{app:post-training-objectives} gives the objective functions
and token-weighting notation.

\subsection{Composability of Task Vectors}

\paragraph{Task Vectors and Merging.}
For parameter-aligned models sharing an initialization $\theta_0$,
let $\theta_t^m$ denote the parameters after training on task $t$
using $m\in\{\mathrm{OPD},\mathrm{RLVR}\}$.
The corresponding task vector is the parameter displacement from the
shared initialization \citep{ilharco2023taskarithmetic}:
\begin{equation}
\tau_t^m=\theta_t^m-\theta_0.
\label{eq:updates}
\end{equation}
In particular, an OPD task vector is measured from the student's
initialization rather than from its teacher.
For a collection of task vectors $\mathcal V$, we write a general
composition as
\begin{equation}
\theta_{\mathrm{comp}}
=
\theta_0+\mathcal F_\phi(\mathcal V),
\label{eq:general-composition}
\end{equation}
where $\mathcal F_\phi$ denotes a merging operator with configuration
$\phi$.
Task Arithmetic (TA) uses a scaled sum,
$\mathcal F_\lambda^{\mathrm{TA}}(\mathcal V)
=\lambda\sum_{\tau\in\mathcal V}\tau$
\citep{ilharco2023taskarithmetic}.
This formulation also includes TIES-Merging
\citep{yadav2023ties} and TSV-M \citep{gargiulo2025tsv},
which transform task vectors before aggregation.

\paragraph{Composability of OPD Task Vectors.}
We use \emph{composability} to describe how effectively task-specific
capabilities are retained or improved when multiple task vectors are
combined.
Our experiments consider three composition settings.
For a task $t$, \textbf{within-task composition} combines its OPD and
RLVR vectors,
\begin{equation}
\mathcal V_t^{\mathrm{within}}
=
\{\tau_t^{\mathrm{OPD}},\tau_t^{\mathrm{RLVR}}\}.
\end{equation}
For a set of tasks $\mathcal T$, \textbf{cross-task composition}
combines vectors produced by the same post-training method,
$\mathcal V_m^{\mathrm{cross}}
=
\{\tau_t^m:t\in\mathcal T\}$.
Finally, \textbf{joint composition} combines vectors across both tasks
and post-training methods,
$\mathcal V^{\mathrm{joint}}
=
\{\tau_t^m:t\in\mathcal T,\,
m\in\{\mathrm{OPD},\mathrm{RLVR}\}\}$.
We evaluate these settings under multiple merging operators, with
training and evaluation details provided in
Section~\ref{sec:setup}.

\section{Composability of OPD Task Vectors}
\label{sec:findings}

We evaluate OPD task vectors in the three composition settings defined
in Section~\ref{sec:preliminaries}. After introducing the experimental
setup (Section~\ref{sec:setup}), we first test whether an OPD update
can improve its own RL teacher within the same task
(Section~\ref{sec:opd-within-task}), then compare OPD and RL updates
when composed across tasks
(Section~\ref{sec:opd-cross-task-composability}). Finally, we combine
both sources across tasks to test whether their within-task
complementarity can benefit a unified multi-task model
(Section~\ref{sec:joint_multi_task_composition}).

\subsection{Experimental Setup}
\label{sec:setup}
\label{sec:datasets-evaluation}

\paragraph{Models and training.}
We use SmolLM3-3B with a MixSFT anchor and Qwen3-4B-Instruct-2507
with its instruction-tuned anchor. SmolLM3's Math, Code, and IF RL
experts come from Open-MOPD \citep{gao2026openmopd}; Qwen's five-domain
RL experts come from \citet{wu2026consolidating} and additionally
cover Science and Agent.
For each domain, one OPD student starts from the shared anchor and
uses the corresponding RL expert as its sole teacher. Within each
family and domain, RL and OPD use identical training prompts.
SmolLM3 uses student-top-$k$ feedback and Qwen uses sampled-token
feedback (Appendix~\ref{app:post-training-objectives}).

\paragraph{Evaluation.}
Both model families are evaluated on AIME 2024 and 2025 for Math.
The remaining benchmarks follow the expert releases:
LiveCodeBench for Code, IFEval and IFBench for IF, and, for Qwen only,
GPQA-Diamond for Science and a BFCL v3 subset for Agent
\citep{gao2026openmopd,wu2026consolidating,llmfusionrepo2026}.
All checkpoints we evaluate within a family--benchmark pair use identical decoding
and evaluation settings. We average benchmarks within domains and
then weight domains equally; the two families' averages cover
different task sets. Our experimental scores are means over three repetitions.
Here avg@$k$ is average correctness over $k$
independent responses, not pass@$k$. Training datasets, benchmark
versions, subset sizes, and sampling counts are listed in
Appendix~\ref{app:training-datasets}--\ref{app:evaluation-datasets}.

\paragraph{Merging.}
We compare TA, TIES, and TSV-M (Section~\ref{sec:preliminaries}),
including Raw Sum as TA with $\lambda=1$.
Each merge combines task vectors from one model family and anchor. Each multi-task
row uses one materialized merged checkpoint across all domains.

\subsection{Within-Task Complementarity of OPD Task Vectors}
\label{sec:opd-within-task}

\begin{table}[!htbp]
    \centering
    \normalsize
    \setlength{\tabcolsep}{4pt}
    \renewcommand{\arraystretch}{1.0}
    \caption{Within-task composition on \textsc{Qwen3-4B} (\%).}
    \label{tab:opd-within-task}

    \begin{tabularx}{\linewidth}{@{}l*{3}{>{\centering\arraybackslash}X}@{}}
        \toprule
        Method & Math & Code & IF \\
        \midrule
        \textcolor{TblRLInk}{RL}
        & 60.14 & 37.51 & 69.53 \\
        \textcolor{TblOPDInk}{OPD}
        & 59.58 & 37.18 & 68.60 \\
        \midrule
        TA
        & \ScoreWithPosDelta{60.51}{0.37}
        & \ScoreWithNegDelta{37.48}{0.03}
        & \ScoreWithNegDelta{68.92}{0.61} \\
        TIES
        & \ScoreWithPosDelta{60.35}{0.21}
        & \ScoreWithPosDelta{41.24}{3.73}
        & \ScoreWithPosDelta{69.66}{0.13} \\
        TSV-M
        & \ScoreWithNegDelta{56.51}{3.63}
        & \ScoreWithPosDelta{38.52}{1.01}
        & \ScoreWithNegDelta{69.29}{0.24} \\
        Raw Sum
        & \ScoreWithNegDelta{59.78}{0.36}
        & \ScoreWithPosDelta{41.01}{3.50}
        & \ScoreWithPosDelta{70.83}{1.30} \\
        \bottomrule
    \end{tabularx}

    \par\smallskip
    {\normalsize\raggedright
    Parentheses show percentage-point changes relative to the RL teacher.\par}
\end{table}

For each of the three Qwen domains shown in Table~\ref{tab:opd-within-task},
we merge the OPD task vector with the task vector of its corresponding
RL teacher using TA, TIES, TSV-M, or Raw Sum. Since the RL teacher
consistently outperforms the standalone OPD expert, we assess
complementarity by whether the merged model surpasses the RL teacher,
which necessarily implies outperforming both constituent models.

Overall, seven of the 12 task--merging configurations exceed both
endpoints in point estimate. The largest gains occur on Code: TIES
and Raw Sum improve over the RL teacher by $3.73$ and $3.50$
percentage points (pp), respectively, while TSV-M yields a $1.01$ pp
gain. Raw Sum also improves instruction following by $1.30$ pp.
On Math, TA achieves the highest score, exceeding the RL teacher by
$0.37$ pp, followed by TIES with a $0.21$ pp gain.

The benefit of composition depends on both the task and the merging
rule. TIES improves over the RL teacher on Math, Code, and IF, whereas
Raw Sum does so on Code and IF. These results show that, despite
its weaker standalone performance, an OPD task vector can provide
complementary updates that produce a model stronger than either
constituent, although such complementarity is not consistently
recovered by every merging configuration.

\finding{finding:within}
{Within a task, an OPD task vector can improve its RL teacher even when the OPD expert has weaker standalone performance.}

\subsection{Cross-Task Composability of OPD Task Vectors}
\label{sec:opd-cross-task-composability}

For the cross-task setting in Table~\ref{tab:opd-cross-task-results},
we merge task vectors from a single training source, either OPD or
RL, into one model using Raw Sum, TA, TIES, or TSV-M. We evaluate
three domains on SmolLM3-3B and five on Qwen3-4B. Since the OPD and
RL experts differ in standalone performance, we assess composition
using both the merged model's equal-weight domain average and its
change from the corresponding standalone specialist average. This
change measures how much performance is gained or lost when separate
experts are consolidated into one model.

Overall, OPD compositions achieve higher averages than their RL
counterparts in seven of the eight backbone--merging-rule comparisons,
despite starting from weaker standalone experts. Under TIES, OPD
reaches $32.71\%$ on SmolLM3-3B and $56.19\%$ on Qwen3-4B, compared
with $32.16\%$ and $55.38\%$ for RL, respectively. The exception is
Raw Sum on SmolLM3-3B, where RL reaches $32.45\%$ and OPD reaches
$32.06\%$. Thus, the final merged scores generally favor OPD, although
the advantage depends on the backbone and merging rule.

Relative to their own specialists, OPD compositions show a larger
average gain or a smaller average loss in all eight comparisons.
For example, TIES on SmolLM3-3B improves the OPD specialist average
by $2.37$ pp, compared with $0.26$ pp for RL. On Qwen3-4B, OPD
approximately preserves its specialist average ($+0.03$ pp), whereas
RL loses $1.85$ pp. These results indicate that OPD updates retain
specialist performance more effectively on average when combined
across tasks. The benefit is not uniform across individual domains:
under Raw Sum, the Qwen RL composition still scores higher on Code
and IF.

\begin{table}[t]
    \centering
    \setlength{\belowcaptionskip}{\baselineskip}
    \caption{Cross-task composition of \TblOPD{} and \TblRL{}
    task vectors (\%). Avg.: equal-weight domain mean.
    Bold marks the higher displayed score in each merging pair
    (both for ties), not statistical significance.
    Single denotes separate task-specific experts.}
    \label{tab:opd-cross-task-results}

    \TblBody
    \setlength{\tabcolsep}{2.5pt}
    \begin{tabular*}{\linewidth}{@{\extracolsep{\fill}}cc*{10}{r}@{}}
        \toprule
        & & \multicolumn{4}{c}{\textsc{SmolLM3-3B}}
          & \multicolumn{6}{c}{\textsc{Qwen3-4B}} \\
        \cmidrule(lr){3-6}\cmidrule(l){7-12}
        Method & Src.
        & \multicolumn{1}{c}{Math} & \multicolumn{1}{c}{Code}
        & \multicolumn{1}{c}{IF} & \multicolumn{1}{c}{Avg.}
        & \multicolumn{1}{c}{Math} & \multicolumn{1}{c}{Code}
        & \multicolumn{1}{c}{IF} & \multicolumn{1}{c}{Sci.}
        & \multicolumn{1}{c}{Agent} & \multicolumn{1}{c}{Avg.} \\
        \midrule
        \multirow{2}{*}{Single} & \TblRL
        & 24.38 & 23.08 & 48.23 & 31.90
        & 60.14 & 37.51 & 69.53 & 44.95 & 74.00 & 57.23 \\
         & \TblOPD
        & 20.42 & 22.69 & 47.92 & 30.34
        & 59.58 & 37.18 & 68.60 & 42.42 & 73.00 & 56.16 \\
        \midrule
        \multirow{2}{*}{Raw Sum} & \TblRL
        & \TblRLWin{25.21} & 23.39 & \TblRLWin{48.75} & \TblRLWin{32.45}
        & 58.19 & \TblRLWin{40.97} & \TblRLWin{69.51} & 35.86 & 69.50 & 54.81 \\
         & \TblOPD
        & 24.79 & \TblOPDWin{24.24} & 47.14 & 32.06
        & \TblOPDWin{60.56} & 40.04 & 68.71 & \TblOPDWin{38.89} & \TblOPDWin{70.00} & \TblOPDWin{55.64} \\
        \addlinespace[1.5pt]
        \multirow{2}{*}{TA} & \TblRL
        & 18.31 & 18.23 & \TblRLWin{44.14} & 26.89
        & 53.61 & 34.87 & \TblRLWin{58.18} & 42.93 & \TblRLWin{64.50} & 50.82 \\
         & \TblOPD
        & \TblOPDWin{19.12} & \TblOPDWin{18.98} & 44.01 & \TblOPDWin{27.37}
        & \TblOPDWin{54.86} & \TblOPDWin{35.36} & 57.79 & \TblOPDWin{43.94} & \TblOPDWin{64.50} & \TblOPDWin{51.29} \\
        \addlinespace[1.5pt]
        \multirow{2}{*}{TIES} & \TblRL
        & 25.63 & \TblRLWin{24.50} & 46.34 & 32.16
        & 57.78 & 41.23 & 69.03 & 37.37 & 71.50 & 55.38 \\
         & \TblOPD
        & \TblOPDWin{26.46} & 23.36 & \TblOPDWin{48.32} & \TblOPDWin{32.71}
        & \TblOPDWin{58.06} & \TblOPDWin{41.32} & \TblOPDWin{69.18} & \TblOPDWin{40.40} & \TblOPDWin{72.00} & \TblOPDWin{56.19} \\
        \addlinespace[1.5pt]
        \multirow{2}{*}{TSV-M} & \TblRL
        & 19.17 & 19.89 & 43.71 & 27.59
        & 56.94 & \TblRLWin{37.22} & \TblRLWin{64.46} & 39.39 & 68.00 & 53.20 \\
         & \TblOPD
        & \TblOPDWin{20.21} & \TblOPDWin{20.64} & \TblOPDWin{43.96} & \TblOPDWin{28.27}
        & \TblOPDWin{57.64} & 36.42 & 62.90 & \TblOPDWin{43.94} & \TblOPDWin{70.00} & \TblOPDWin{54.18} \\
        \bottomrule
    \end{tabular*}
    \par\smallskip
\end{table}

\finding{finding:cross}
{Cross tasks, OPD task vectors can form stronger multi-task models and preserve specialist performance more effectively on average than RL task vectors.}

\subsection{OPD Task Vectors in Joint Multi-Task Composition}
\label{sec:joint_multi_task_composition}

We next examine whether the within-task complementarity between OPD
and RL updates can improve a model that combines multiple tasks.
Joint composition introduces many possible choices of within-task
source weights, cross-task weights, and merging operators, making an
exhaustive search costly. The preceding results motivate a simple
construction: within-task combinations can improve the RL teacher,
while cross-task Raw Sum outperforms scaled TA on both backbones.
We therefore average the OPD and RL updates within each task and
directly sum the resulting updates across tasks, using the same
recipe for every domain.

Specifically, we construct
$\theta_{\mathrm{joint}}=\theta_0+\sum_{t=1}^{K}(o_t+r_t)/2$,
where $o_t$ and $r_t$ denote the OPD and RL task vectors for task $t$,
and $K=3$ for SmolLM3-3B and $K=5$ for Qwen3-4B. Averaging assigns
equal weight to the two sources and keeps the sum of their
coefficients within each task at one, matching a single-source Raw
Sum. Table~\ref{tab:joint_composition_backbones} compares this joint
construction with compositions built from OPD or RL alone; each
result corresponds to one checkpoint evaluated across all domains.

\begin{table}[t]
    \centering
    \normalsize
    \setlength{\tabcolsep}{1.2pt}
    \renewcommand{\arraystretch}{1.08}
    \caption{\textbf{Joint composition of OPD and RL task vectors across backbones.}
    Scores are percentages; Avg. equally weights the domains available for each backbone. Bold marks column-wise maxima within each backbone.}
    \label{tab:joint_composition_backbones}

    \begin{tabularx}{\linewidth}{@{}
        >{\centering\arraybackslash}p{0.12\linewidth}
        *{4}{>{\centering\arraybackslash}X}
        @{\hspace{4pt}}
        *{6}{>{\centering\arraybackslash}X}@{}}
        \toprule
        & \multicolumn{4}{c}{\textsc{SmolLM3-3B}}
        & \multicolumn{6}{c}{\textsc{Qwen3-4B}} \\
        \cmidrule(lr){2-5}\cmidrule(l){6-11}
        Method
        & Math & Code & IF & Avg.
        & Math & Code & IF & Sci. & Agent & Avg. \\
        \midrule
        \GtwoRL
        & 25.21 & 23.39 & 48.75 & 32.45
        & 58.19 & \textbf{40.97} & \textbf{69.51} & 35.86 & 69.50 & 54.81 \\
        \GtwoOPD
        & 24.79 & 24.24 & 47.14 & 32.06
        & \textbf{60.56} & 40.04 & 68.71 & 38.89 & 70.00 & 55.64 \\
        \TblOPD + \TblRL
        & \textbf{26.04} & \textbf{24.64} & \textbf{49.16} & \textbf{33.28}
        & 60.42 & 40.39 & 68.64 & \textbf{40.40} & \textbf{73.50} & \textbf{56.67} \\
        \bottomrule
        
    \end{tabularx}
    \vspace{-10pt}
\end{table}

On Qwen3-4B, joint composition achieves an average of $56.67\%$,
compared with $54.81\%$ for RL-only and $55.64\%$ for OPD-only
composition. Relative to RL-only composition, it improves Math,
Science, and Agent, while Code and IF decline. The way the sources
are combined also matters: alternative ten-vector Raw Sum and TIES
configurations score $52.16\%$ and $48.10\%$, respectively. The
balanced construction therefore achieves the highest observed
average among these configurations. These comparisons use means over
three repetitions. The mean improvements are descriptive point
estimates and do not, by themselves, establish statistical significance.

\begingroup
\setlength{\columnsep}{10pt}
\setlength{\intextsep}{\baselineskip}
\begin{wraptable}{r}{0.50\textwidth}
    \centering
    \normalsize
    \setlength{\tabcolsep}{1.2pt}
    \renewcommand{\arraystretch}{1.08}
    \caption{\textbf{Comparison with multi-model OPD baselines on \textsc{SmolLM3-3B}.}
    Scores are percentages; Avg. equally weights Math, Code, and IF. Bold marks column-wise maxima.}
    \label{tab:multi_model_opd_smol}

    \begin{tabularx}{\linewidth}{@{}
        >{\centering\arraybackslash}p{0.31\linewidth}
        *{4}{>{\centering\arraybackslash}X}@{}}
        \toprule
        Method & Math & Code & IF & Avg. \\
        \midrule
        \TblRL
            & 25.21 & 23.39 & 48.75 & 32.45 \\
        \TblOPD
            & 24.79 & 24.24 & 47.14 & 32.06 \\
        \TblOPD + \TblRL
            & \textbf{26.04} & \textbf{24.64} & 49.16 & \textbf{33.28} \\
        \midrule
        Naive M-OPD
            & 21.26 & 19.26 & 43.64 & 28.05 \\
        Open-MOPD
            & 22.42 & 21.73 & \textbf{49.58} & 31.24 \\
        \bottomrule
    \end{tabularx}

\end{wraptable}

On SmolLM3-3B, joint composition reaches $33.28\%$, exceeding
$32.45\%$ for RL-only and $32.06\%$ for OPD-only composition.
It also surpasses both single-source compositions in each domain,
with scores of $26.04\%$ on Math, $24.64\%$ on Code, and $49.16\%$
on IF. The benefit therefore extends across all three evaluated
domains on SmolLM3, while the Qwen results show task-dependent
trade-offs.

Table~\ref{tab:multi_model_opd_smol} additionally compares the
SmolLM3 results with published multi-teacher distillation baselines
from \citet{gao2026openmopd}. Joint composition has the highest
displayed average, while Open-MOPD has the highest IF score
($49.58\%$) and an average of $31.24\%$. These published results
provide useful reference points, but they come from separate runs
and use Math avg@64 rather than our avg@8, limiting conclusions
about relative method effectiveness.

\par
\endgroup

\finding{finding:joint}
{Joint composition by simple averaging and summation yields higher observed multi-task mean scores than either single-source composition on both evaluated backbones.}

\section{Understanding the Task Vectors from OPD}
\label{sec:analysis}

In this section, we examine update structure, within-task complementarity, and cross-task geometry.

\subsection{Parameter Structure of OPD Task Vectors}
\label{sec:opd-geometry}

\begin{figure}[!ht]
    \centering
    \includegraphics[width=0.2327\linewidth]{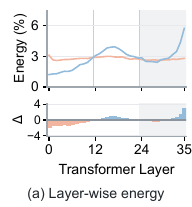}\hfill
    \includegraphics[width=0.2327\linewidth]{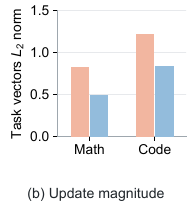}\hfill
    \includegraphics[width=0.2327\linewidth]{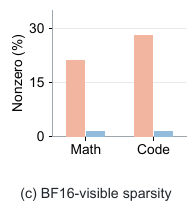}\hfill
    \includegraphics[width=0.2327\linewidth]{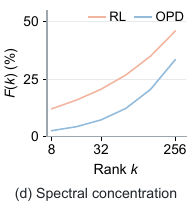}
    \caption{\textsc{SmolLM3-3B} \textbf{task-vector structure.}
    \textbf{(a)} Code layer-wise energy share; shading marks the final 12 layers
    and $\Delta$ denotes OPD minus RL (pp).
    \textbf{(b)} Update $L_2$ norms and \textbf{(c)} BF16-visible nonzero fractions
    for Math and Code. \textbf{(d)} Code rank-$k$ energy retention $F(k)$.
    Full profiles: Fig.~\ref{fig:smollm-structure-full}.}
    \label{fig:smollm-structure}
    \vspace{-10pt}
\end{figure}

Figure~\ref{fig:smollm-structure} compares Math and Code task vectors
from a common SmolLM3-3B base. OPD's update norm is $59.7\%$ of RL's
for Math and $68.7\%$ for Code. OPD changes fewer saved BF16 coordinates
and has lower spectral concentration at all six measured ranks from
8 to 256. The final 12 layers contain $37.7\%$ versus $32.1\%$ of
each vector's squared $L_2$ energy for Code, and $44.7\%$ versus
$32.3\%$ for Math. These larger relative shares coexist with lower
absolute late-layer energy.

Spectral concentration $F(k)$ aggregates rank-$k$ retained energy over
252 attention and MLP projection matrices, using approximate randomized
SVD and normalizing by their total update energy. Coordinate sparsity
and spectral concentration capture different properties: fewer changed
coordinates can coexist with energy spread across singular directions.
Appendix~\ref{app:parameter-measurements} gives complete measurements and
definitions.

\subsection{Within-Task Parameter Complementarity between OPD and RL}
\label{sec:within-task-complementarity}

\begin{figure}[!ht]
    \centering
    \includegraphics[width=\linewidth]{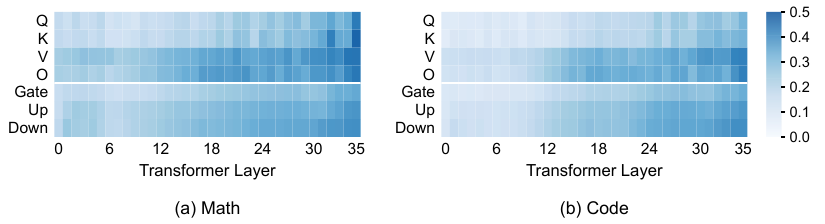}
    \caption{\textbf{Layer-wise RL--OPD alignment on }\textsc{SmolLM3-3B.}
    Task-vector matrix cosine similarities for \textbf{(a)} Math and
    \textbf{(b)} Code, with a shared color scale. Columns index Transformer
    layers; rows index attention (Q, K, V, O) and MLP (Gate, Up, Down)
    projections.}
    \label{fig:within-task-local-alignment}
    \vspace{-10pt}
\end{figure}

We ask whether combining RL and OPD updates provides a useful direction
beyond matching update magnitude. All analyses in this subsection use
SmolLM3-3B. The Code controls reuse the RL--OPD checkpoint pair used for
the Code geometry measurements. Figure~\ref{fig:within-task-local-alignment}
shows that RL--OPD alignment varies across layers and projections.

For RL and OPD task vectors $r$ and $o$, write
$o=\alpha r+q$, where $\alpha=\langle o,r\rangle/\norm{r}_2^2$
and $q\perp r$. The orthogonal energy share,
$\norm{q}_2^2/\norm{o}_2^2$, is $93.0\%$ for Math and $95.2\%$
for Code in the measured task vectors. These values indicate substantial
non-collinearity, but do not establish whether the residual improves
task performance.

To test directional complementarity on Code, we set
$R=\norm{r+o}_2$ and rescale the RL-only and OPD-only updates to this
norm. Table~\ref{tab:within-task-controls} gives the constructions and
score differences. Raw Sum exceeds the norm-matched RL and OPD controls
by $1.77$ and $2.07$ pp, respectively. Thus, at the tested global update
norm, the combined direction yields higher mean Code scores than either
constituent direction.

\begin{table}[!ht]
    \centering
    \caption{\textbf{Within-task composition on }\textsc{SmolLM3-3B} \textbf{in} \textsc{Code}: norm-matched controls
    and residual ablation.} All models take the form $\theta_0+u$,
    with $R=\norm{r+o}_2$.
    \label{tab:within-task-controls}
    \normalsize
    \setlength{\tabcolsep}{5pt}
    \renewcommand{\arraystretch}{1.16}
    \begin{tabularx}{\linewidth}{@{}l>{\centering\arraybackslash}Xcc@{}}
        \toprule
        Configuration & Update $u$ & $\norm{u}_2/R$
                      & $\Delta$ vs. Raw Sum (pp) \\
        \midrule
        Raw Sum & $r+o$ & $1$ & $0.00$ \\
        Norm-matched RL & $(R/\norm{r}_2)r$ & $1$ & $-1.77$ \\
        Norm-matched OPD & $(R/\norm{o}_2)o$ & $1$ & $-2.07$ \\
        \midrule
        Residual removed & $(1+\alpha)r$ & $<1$ & $-2.88$ \\
        \bottomrule
    \end{tabularx}
    \par\smallskip
    \normalsize\raggedright
    $\Delta=\mathrm{Score}(\theta_0+u)-\mathrm{Score}(\theta_0+r+o)$,
    using the reported mean scores. The last row is not norm-matched:
    removing $q$ reduces both the score and update norm. The Code geometry
    measurements and controls in this subsection use the same checkpoint pair
    (Appendix~\ref{app:within-task-diagnostics}).\par
\end{table}

\Needspace{6\baselineskip}
Removing $q$ from $r+o=(1+\alpha)r+q$ lowers the mean score by $2.88$ pp.
Since $q\perp r$, this also reduces the update norm and does not by itself
isolate a directional effect. The norm-matched comparison supports
directional complementarity in the tested configuration; it does not
establish superiority over the best scalar rescaling of either
constituent update. Appendix~\ref{app:within-task-diagnostics} provides the
measurement scope and a local loss interpretation of the residual.

\subsection{How Update Structure Shapes Multi-Task Composition}
\label{sec:multitask-mechanism}

On Qwen3-4B, we examine how updates from other tasks enter a target
task's high-energy channels and alter its direction during additive
composition.

\begin{figure}[!ht]
    \centering
    \includegraphics[width=\linewidth]{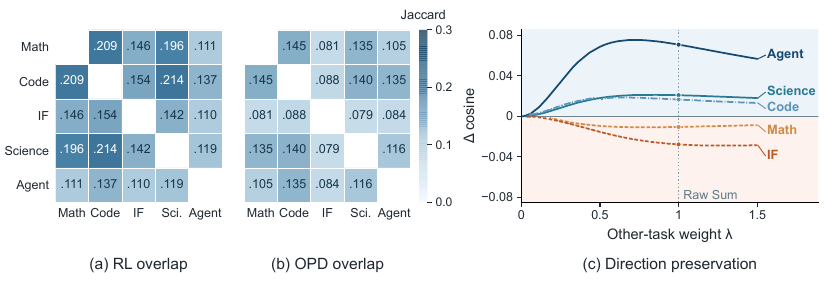}
    \caption{\textbf{Cross-task overlap and direction retention on }\textsc{Qwen3-4B}.
    \textbf{(a,b)} Jaccard overlap of RL/OPD top-$10\%$ MLP-channel sets
    (shared scale). \textbf{(c)} OPD--RL difference in
    $c_t(\lambda)=\cos(u_t,u_t+\lambda v_t)$; positive values favor OPD,
    $\lambda=1$ denotes Raw Sum.}
    \label{fig:multitask-geometry}
\end{figure}

For each layer, we select the top $10\%$ of MLP channels by squared
update magnitude. OPD exhibits lower Jaccard overlap than RL across all
ten task pairs, with the mean decreasing from $0.15368$ to $0.11064$,
a $28\%$ reduction (Fig.~\ref{fig:multitask-geometry}a,b). Thus, the
high-energy regions of OPD updates are less shared across tasks.

Let $S_t$ denote task $t$'s hotspots and $\tau_{s,S_t}$ the update from
task $s$ restricted to them. The summed individual off-task energy
relative to the target task's own energy is
\begin{equation}
R_t=\frac{\sum_{s\neq t}\norm{\tau_{s,S_t}}_2^2}
{\norm{\tau_{t,S_t}}_2^2}.
\label{eq:relative-incoming-energy}
\end{equation}
OPD has lower $R_t$ than RL on Code, Science, and Agent, and
higher $R_t$ on Math and IF. For Science, $R_t$ decreases from $3.613$
to $3.262$; for Agent, it decreases from $4.637$ to $3.103$.
Thus, relative incoming energy varies by task despite the overall
reduction in hotspot overlap. Appendix~\ref{app:multitask-diagnostics}
decomposes $R_t$ into relative update scale and localization.

We next measure how these updates align and accumulate. For
$u_t=\tau_{t,S_t}$ and $v_t=\sum_{s\neq t}\tau_{s,S_t}$, direction
retention is
\begin{equation}
c_t(\lambda)=\cos\!\left(u_t,\;u_t+\lambda v_t\right),
\label{eq:hotspot-direction-retention}
\end{equation}
where $\lambda$ controls the weight of the other-task updates. At
$\lambda=1$, corresponding to Raw Sum, OPD improves direction retention
by $0.016$, $0.021$, and $0.071$ for Code, Science, and Agent,
respectively. Direction retention decreases by $0.011$ for Math and
$0.028$ for IF (Fig.~\ref{fig:multitask-geometry}c).

Across the five tasks, lower relative incoming energy coincides with
better direction retention for Code, Science, and Agent, while Math
and IF show the opposite pattern. These observations connect update
distribution and relative magnitude with direction retention in the
measured hotspots.

\takeaway{OPD contributes update directions beyond those of its RL
teachers and exhibits lower overlap across tasks. These properties
provide a geometric perspective on its within-task complementarity and
cross-task composability.}

\section{Related Work}
\label{sec:related}

\paragraph{On-Policy Distillation and Capability Integration.}
GKD studies flexible teacher--student divergence objectives for
student-generated trajectories, while MiniLLM develops reverse-KL
optimization \citep{agarwal2024gkd,gu2024minillm}.
MOPD and Open-MOPD study multi-teacher capability integration during
training \citep{ma2026mopd,gao2026openmopd}.
We instead study whether the resulting task-specific OPD updates can
compose in parameter space. Further details appear in
Appendix~\ref{app:opd-related-work}.

\paragraph{Task Vectors and Model Merging.}
Model Soups~\citep{wortsman2022soups} can improve accuracy and robustness
by averaging models fine-tuned from a common initialization with different
hyperparameters, without increasing inference cost.
Task Arithmetic~\citep{ilharco2023taskarithmetic} defines task vectors as
fine-tuned minus initial weights and composes or edits models through
arithmetic on these vectors. TIES-Merging~\citep{yadav2023ties} trims
small-magnitude changes and resolves sign conflicts;
DARE~\citep{yu2024dare} randomly drops and rescales parameter deltas.
These methods motivate our study of OPD-derived task-vector composition.

\paragraph{Geometry and Composability of Post-Training Updates.}
Task arithmetic is linked to weight disentanglement, with parameter
directions inducing localized function-space changes
\citep{ortizjimenez2023tangent}. LLM post-training studies examine SFT--RLVR
structural differences for capability synthesis~\citep{yuan2026dots}
and relate update sparsity and cross-domain alignment to merging
behavior~\citep{wu2026sparsitycurse,wu2026consolidating}.
OPD studies identify checkpoint-precision coordinate sparsity, spectral
concentration, and early subspace locking, and demonstrate effective
training in restricted update subspaces or
subnetworks~\citep{shen2026opdgeometry,yu2026densesupervision}.
We examine how OPD update geometry relates to composition with RL
updates within a task and with OPD updates across tasks.
% END SOURCE: sections/related.tex

% BEGIN SOURCE: sections/discussion.tex
\section{Discussion and Conclusion}
\label{sec:discussion}

We study the composability of task vectors produced by on-policy distillation (OPD), using reinforcement-learning (RL) teacher updates as the main comparison. Across five domains and two model architectures, we find that \textbf{within a task}, merging OPD and RL task vectors can outperform both constituent models even when the OPD student is weaker, while \textbf{across tasks}, OPD compositions achieve higher average performance than corresponding RL compositions in seven of eight backbone--merging-rule comparisons. OPD updates are not collinear with RL updates, and SmolLM3-3B Code controls support directional complementarity at the tested global update norm. These results suggest that task-vector composability captures a distinct property of post-training updates beyond standalone model performance.

\paragraph{Limitations and Future Work.}
We evaluate general-purpose merging rules. Future work could exploit
the observed differences in update direction, scale, and concentration
to develop OPD-aware merging methods and evaluate them across additional
tasks and model families.
\label{maintext:end}
% Keep declarations and references outside the main text.

\section*{Reproducibility statement}
Section~\ref{sec:preliminaries} defines the objects, and
Section~\ref{sec:datasets-evaluation} introduces the evaluation protocols;
Sections~\ref{sec:opd-within-task}--\ref{sec:joint_multi_task_composition} specify the
merging experiments. Section~\ref{sec:analysis} presents the associated
analyses and controls. Appendix~\ref{app:experimental-details}
documents the data, OPD training configurations, merging
parameters, and score aggregation. All experimental results obtained
in this study are supported by statistical analyses. Performance
scores are reported as arithmetic means over three repetitions.

\section*{AI use statement}
Generative AI assisted with language polishing.

\bibliography{references}
\bibliographystyle{iclr2027_conference}
\clearpage
\appendix
% BEGIN SOURCE: sections/appendix.tex

\section{Post-Training Objectives}
\label{app:post-training-objectives}

Post-training adapts pretrained models through supervised fine-tuning
(SFT), reinforcement learning, and distillation. SFT learns from
demonstrations and can provide a cold-start initialization for
subsequent optimization \citep{deepseekai2025r1}. We focus on RLVR
and policy-gradient OPD.

Let $\pi_\theta$ denote the policy being optimized,
$\pi_{\mathrm{old}}$ the behavior policy used to generate the current
rollouts, and $y\sim\pi_{\mathrm{old}}(\cdot\mid x)$ a response to
$x\sim\mathcal D$, with $h_i=(x,y_{<i})$.

\paragraph{Reinforcement Learning with Verifiable Rewards (RLVR).}
RLVR constructs advantages $\widehat A_i^{\mathrm{RLVR}}$ from
verifier rewards $R(x,y)$ obtained through answer checks or executable
tests \citep{deepseekai2025r1}. Omitting clipping, regularization,
and loss normalization, its policy-gradient surrogate is
\citep{shao2024deepseekmath}
\begin{equation}
\mathcal J_{\mathrm{RLVR}}(\theta)=
\mathbb E_{\substack{x\sim\mathcal D\\
y\sim\pi_{\mathrm{old}}(\cdot\mid x)}}
\left[
\sum_{i=1}^{|y|}
\rho_i(y_i)
\operatorname{sg}\!\left(\widehat A_i^{\mathrm{RLVR}}\right)
\right],
\label{eq:policy-gradient}
\end{equation}
where
$\rho_i(v)=\pi_\theta(v\mid h_i)/\pi_{\mathrm{old}}(v\mid h_i)$
and $\operatorname{sg}$ denotes stop-gradient.
In outcome-based RLVR, the advantages are derived from
response-level verification.

\paragraph{On-Policy Distillation (OPD).}
Following the supervision-granularity distinction of
\citet{li2026rethinking}, we consider sampled-token and student-top-$k$
feedback. As in Open-MOPD \citep{gao2026openmopd}, a fixed teacher $q$
provides log-probability feedback on student-generated prefixes.
We write the token-feedback signal used here as
\begin{equation}
\widehat A_i^{\mathrm{OPD}}(v)
=
w_i(v)
\left[
\log q(v\mid h_i)
-
\log\pi_{\mathrm{old}}(v\mid h_i)
\right],
\qquad v\in S_i,
\label{eq:opd-token-signal}
\end{equation}
where $S_i$ and $w_i$ specify the supervised tokens and their weights.
Sampled-token OPD uses $S_i=\{y_i\}$ and $w_i=1$, while top-$k$ OPD
supervises multiple selected candidates.
The corresponding policy-gradient surrogate aggregates
$\rho_i(v)\operatorname{sg}(\widehat A_i^{\mathrm{OPD}}(v))$
over $v\in S_i$ at each prefix.
Thus, RLVR and OPD share a closely related policy-gradient structure
but differ in the source and granularity of their learning signals:
RLVR optimizes verifier-defined outcomes, whereas OPD aligns the
student toward a teacher through token-level feedback.

\clearpage
\section{Detailed Experimental Setup}
\label{app:experimental-details}

\paragraph{Models and Experts.}
We study two model families: SmolLM3-3B and
Qwen3-4B-Instruct-2507.
Their task-vector anchors are the SmolLM3 MixSFT checkpoint and
Qwen3-4B-Instruct-2507, respectively.
For SmolLM3-3B, we use the three domain-specific RL experts released
by Open-MOPD, covering \textbf{mathematics}, \textbf{coding}, and
\textbf{instruction following}
\citep{gao2026openmopd}.
For Qwen3-4B-Instruct-2507, we use the five-domain RL experts released
by \citet{wu2026consolidating}, which additionally cover
\textbf{science} and \textbf{agentic tool use}.
For each domain, we train one OPD student from the same task-vector
anchor, using the corresponding RL expert as its sole teacher.
Both model families follow the policy-gradient OPD formulation in
Appendix~\ref{app:post-training-objectives}; the SmolLM3 implementation uses
student-top-$k$ feedback, whereas the Qwen implementation uses
sampled-token feedback.

\paragraph{Data provenance of the reused SmolLM3 anchor.}
The released MixSFT checkpoint was trained on three processed SFT
collections: 93,733 OpenR1-Math examples with verified mathematical
reasoning traces \citep{openr12025math}; 50,000 sampled
OpenCodeReasoning examples containing programming problems and
reasoning/code responses \citep{ahmad2025opencodereasoning}; and
820,039 Instruction-Nemotron-aligned examples derived from the
Llama-Nemotron post-training collection
\citep{bercovich2025llamanemotron,nvidia2025llamaposttraining}.
These counts describe Open-MOPD's released SFT configurations
\citep{gao2026openmopd,openmopd2026data}.
We reuse this checkpoint as the anchor; the domain-specific OPD runs
use the prompt sets described below.

\subsection{Training Datasets}
\label{app:training-datasets}

We use the training prompts associated with each family's released RL
experts. Within a family and domain, RL and OPD use the same prompt set;
RL receives the dataset's verifier reward, whereas OPD receives feedback
from the corresponding RL teacher on student-generated responses.
Table~\ref{tab:training-datasets} summarizes the processed training
splits in the Open-MOPD and LLM-Fusion releases
\citep{gao2026openmopd,openmopd2026data,wu2026consolidating,llmfusiontrain2026}.
Counts refer to these processed releases, after their preparation and
filtering, rather than to the full upstream collections.

\begin{table}[!htbp]
    \centering
    \normalsize
    \setlength{\tabcolsep}{5pt}
    \renewcommand{\arraystretch}{1.10}
    \caption{Training-data provenance and released split sizes.
    Each row supplies the prompts for one domain-specific RL expert
    and its corresponding OPD student.}
    \label{tab:training-datasets}
    \begin{tabularx}{\linewidth}{@{}lXr@{}}
        \toprule
        Domain & Source of the processed training prompts & Released examples \\
        \midrule
        \multicolumn{3}{@{}l}{\textbf{SmolLM3-3B: Open-MOPD-Data}} \\
        Math & DAPO-Math-17k & 17,917 \\
        Code & DeepCoder-Preview, excluding LiveCodeBench-source prompts & 23,667 \\
        IF & Nemotron-Cascade-2-RL-data, IF-RL subset & 45,347 \\
        \midrule
        \multicolumn{3}{@{}l}{\textbf{Qwen3-4B: LLM-Fusion-Train}} \\
        Math & Polaris-Dataset-53K & 38,131 \\
        Science & OpenScienceReasoning-2 & 50,000 \\
        Code & Nemotron-RL competitive coding & 19,169 \\
        IF & Nemotron-RL instruction following & 16,575 \\
        Agent & Nemotron-RL workplace assistant & 10,229 \\
        \bottomrule
    \end{tabularx}
\end{table}

\paragraph{SmolLM3 mathematics.}
DAPO-Math-17k contains mathematical problems collected from web and
competition sources, with selection and reformulation to obtain
integer-valued answers suitable for rule-based verification
\citep{yu2025dapo}.
The Open-MOPD preparation reformats the problem prompt and preserves
the reference answer; its released Math portion contains 17,917 examples.

\paragraph{SmolLM3 coding.}
DeepCoder-Preview pairs programming problems with executable test cases
\citep{luo2025deepcoder,deepcoder2025dataset}.
Its upstream training sources include TACO, PrimeIntellect, and
LiveCodeBench. The Open-MOPD release excludes the LiveCodeBench-source
training examples and retains 23,667 processed Code prompts from the
remaining sources. This source exclusion is the decontamination step
associated with the coding set used here. The executable tests define
the correctness signal for the RL expert.

\paragraph{SmolLM3 instruction following.}
The IF prompts originate from the \texttt{IF-RL} subset of
Nemotron-Cascade-2-RL-data
\citep{yang2026nemotroncascade2,nvidia2026cascade2data}.
This subset derives from Nemotron-RL-instruction\_following, which
combines WildChat prompts with automatically checkable instruction
constraints \citep{nvidia2025nemotronif}; Cascade-2 corrects inconsistent
formatting of constraint arguments. Each example specifies a prompt,
instruction identifiers, and their arguments, so success can be checked
programmatically. The upstream IF-RL subset has 45,879 examples, and
the processed Open-MOPD IF portion contains 45,347.

\paragraph{Qwen mathematics and science.}
The Math split derives from
\href{https://huggingface.co/datasets/POLARIS-Project/Polaris-Dataset-53K}{Polaris-Dataset-53K},
a mathematical reasoning collection with reference answers and
model-estimated solve rates \citep{an2025polaris}.
The Science split derives from OpenScienceReasoning-2, which contains
synthetic multiple-choice and open-ended questions with reference
answers and reasoning traces across scientific and academic subjects
\citep{nvidia2025opensciencereasoning2}.
We use the preparation of \citet{wu2026consolidating}.
For Math, the source release uses the provided eight-sample
DeepSeek-R1-Distill-Qwen-7B solve-rate annotations and removes problems with
a success rate greater than $4/8$, yielding 38,131 prompts.
For Science, the source authors estimate difficulty with eight responses
from Qwen3-4B-Instruct-2507, remove problems solved more than $6/8$ of
the time, and sample 50,000 from the remaining problems.
We reuse these released splits and their problem/reference-answer
structure.

\paragraph{Qwen coding, instruction following, and agentic tool use.}
The remaining splits originate from NVIDIA's Nemotron-RL
competitive-coding, instruction-following, and workplace-assistant
datasets
\citep{nvidia2025nemotroncoding,nvidia2025nemotronif,nvidia2025nemotronagent}.
Coding examples pair programming tasks with executable tests.
Instruction-following examples pair requests with verifiable
constraints. Workplace-assistant examples specify tasks, available
tools, and environment information for evaluating tool-mediated
actions. The corresponding processed LLM-Fusion splits contain
19,169, 16,575, and 10,229 examples, respectively.

\paragraph{Released artifacts.}
The SmolLM3 counts are recorded in the
\texttt{rl\_prompt\_mix} manifest of
\href{https://huggingface.co/datasets/BytedTsinghua-SIA/Open-MOPD-Data/tree/9e897efe3257599d4300e2d5ee865a1cc714af87}{Open-MOPD-Data, revision \texttt{9e897efe}}.
For Qwen, the five domains correspond to the \texttt{Math},
\texttt{Science}, \texttt{Code}, \texttt{IF}, and \texttt{Agent}
configurations of
\href{https://huggingface.co/datasets/Siye01/LLM-Fusion-Train/tree/7250d328b3bada207f2167de448e847c6dfff016}{LLM-Fusion-Train, revision \texttt{7250d328}},
each using its \texttt{train} split. These release identifiers specify the processed
data underlying Table~\ref{tab:training-datasets}; the original sources
above describe how the constituent tasks were collected and verified.

\subsection{Evaluation Datasets and Scoring}
\label{app:evaluation-datasets}

Table~\ref{tab:evaluation-datasets} summarizes the benchmark instances
and metrics. Both model families use AIME 2024 and 2025 for Math.
The remaining benchmarks use the processed subsets distributed with
Open-MOPD for SmolLM3 \citep{gao2026openmopd,openmopd2026data}
and LLM-Fusion-Test for Qwen
\citep{wu2026consolidating,llmfusiontest2026}.
The descriptions below identify the original benchmarks and the
particular subsets used for evaluation.

\begin{table}[!htbp]
    \centering
    \normalsize
    \setlength{\tabcolsep}{4pt}
    \renewcommand{\arraystretch}{1.15}
    \caption{Evaluation datasets and per-benchmark metrics.
    Counts are benchmark instances before response sampling and
    averaging across experimental repetitions.}
    \label{tab:evaluation-datasets}
    \begin{tabularx}{\linewidth}{@{}llr>{\raggedright\arraybackslash}Xl@{}}
        \toprule
        Domain & Benchmark & Instances & Metric & Backbones \\
        \midrule
        Math & AIME 2024 & 30 & Answer accuracy, avg@8 & Both \\
             & AIME 2025 & 30 & Answer accuracy, avg@8 & Both \\
        Code & LiveCodeBench v5 & 167 & Execution correctness, avg@10 & Both \\
             & LiveCodeBench v6 & 175 & Execution correctness, avg@10 & Both \\
        IF & IFEval & 541 & Strict prompt-level accuracy & Both \\
           & IFBench (test) & 300 & Strict prompt-level accuracy & Both \\
        Science & GPQA-Diamond & 198 & Multiple-choice accuracy & Qwen \\
        Agent & BFCL v3 subset & 200 & Tool-task success & Qwen \\
        \bottomrule
    \end{tabularx}
\end{table}

\Needspace{5\baselineskip}
\paragraph{Mathematical reasoning: AIME.}
The American Invitational Mathematics Examination (AIME) is organized
by the Mathematical Association of America \citep{maaAIME}.
We evaluate the 2024 and 2025 editions, each combining the 15 problems
in AIME I and the 15 problems in AIME II.
Problems require mathematical reasoning and
have integer answers between 0 and 999. We score the extracted final
answer against the reference answer.
For AIME 2024, both model families use the 30-problem
\texttt{HuggingFaceH4/aime\_2024} dataset \citep{h4aime2024}.
A machine-readable version of AIME 2025 is distributed by math-ai
\citep{mathaiaime2025}.
For both SmolLM3-3B and Qwen3-4B-Instruct-2507, the Math score is the
equal-weight mean of AIME 2024 and AIME 2025 avg@8.
Here avg@$k$ averages correctness over $k$ independently generated
responses per problem. The published Open-MOPD baseline scores quoted
in Table~\ref{tab:multi_model_opd_smol} use their original Math avg@64
protocol \citep{gao2026openmopd}.

\paragraph{Code generation: LiveCodeBench.}
LiveCodeBench collects time-stamped competitive programming problems
from LeetCode, AtCoder, and Codeforces
\citep{jain2024livecodebench}.
We use its code-generation task, where a generated program is checked
against executable test cases. We use the incremental evaluation
subsets packaged in the expert releases: 167 problems for v5 and
175 for v6. In the benchmark's
\href{https://github.com/LiveCodeBench/LiveCodeBench/blob/28fef95ea8c9f7a547c8329f2cd3d32b92c1fa24/README.md#dataset-versions}{official versioning},
these increments correspond to October 2024--January 2025 and
February--April 2025, respectively.
For each problem, avg@10 averages the correctness of ten sampled
programs; the Code score equally averages the two benchmark scores.

\paragraph{Instruction following: IFEval and IFBench.}
IFEval contains 541 prompts spanning 25 types of automatically verifiable
response constraints, such as length, keywords, and output structure
\citep{zhou2023ifeval}.
We use strict prompt-level accuracy: a response succeeds when it
satisfies every instruction attached to its prompt under the strict
checker. IFBench introduces 58 verifiable instruction types across
seven groups, including counts, ratios, words, sentences, formats,
custom constraints, and copying, to evaluate generalization in
instruction following \citep{pyatkin2025ifbench}.
We use its 300-prompt test split with strict prompt-level accuracy.
The IF domain score equally averages IFEval and IFBench accuracies.

\paragraph{Scientific reasoning: GPQA-Diamond.}
GPQA consists of expert-written, graduate-level multiple-choice
questions in biology, physics, and chemistry \citep{rein2023gpqa}.
We use the 198-question Diamond subset, whose selection incorporates
expert agreement and difficulty for nonexpert validators.
Each question has four answer options,
and correctness is determined by the selected option. This benchmark
supplies Qwen's Science score.

\paragraph{Agentic tool use: BFCL v3.}
The Berkeley Function Calling Leaderboard evaluates selecting tools,
providing appropriate arguments, and completing tasks through tool
execution \citep{patil2025bfcl}.
We use the 200-instance \texttt{BFCL\_v3/test} subset distributed
with LLM-Fusion-Test, derived from the \texttt{multi\_turn\_base}
category \citep{llmfusiontest2026}.
Each released instance contains one user request and supports multiple
tool-execution steps. The released evaluator replays the predicted
and reference tool calls from the same initial environment and checks
the resulting tool states and execution responses
\citep{llmfusionrepo2026}.
The percentage of successful instances gives the Agent score.

\paragraph{Evaluation coverage.}
For question-level paired analyses, the five Qwen domains contain
1,641 evaluation instances in total.

\paragraph{Repeated experiments and reported means.}
\label{app:repeated-evaluation}
For the experiments conducted in this study, we report the arithmetic
mean of three experimental repetitions. If $s_{d,b}^{(j)}$ is the score
of benchmark $b$ in domain $d$ for repetition $j$, the reported domain
score and overall score are
\begin{equation}
\bar{s}_d=\frac{1}{3}\sum_{j=1}^{3}
\frac{1}{|\mathcal B_d|}\sum_{b\in\mathcal B_d}s_{d,b}^{(j)},
\qquad
\bar{s}=\frac{1}{|\mathcal D|}\sum_{d\in\mathcal D}\bar{s}_d,
\end{equation}
where $\mathcal B_d$ is the benchmark set for domain $d$ and
$\mathcal D$ is the set of evaluated domains for the backbone.
Within each repetition, avg@$k$ is computed using the stated number
of responses per problem; averaging across three repetitions is a
separate aggregation step. Reported improvements are differences
between these mean scores. We do not interpret a positive mean
difference as establishing statistical significance. Published
baseline scores quoted from prior work retain their original
evaluation and aggregation protocols. Parameter-space diagnostics
describe the analyzed checkpoints and are not accuracy averages.

All checkpoints evaluated in this study within the same model-family--benchmark pair
use identical decoding and evaluation settings.
When reporting aggregate performance, we first average benchmarks
within each domain and then assign equal weight to each domain.
Consequently, the three-domain SmolLM3 and five-domain Qwen averages
summarize different task sets and are not directly comparable in
absolute value.

\subsection{Training Configurations}
\label{app:training-configurations}

\paragraph{OPD training configurations.}
We perform full-parameter OPD using the shared optimization and rollout
settings of the corresponding GRPO experts: the SmolLM3 configurations
in Appendix~8.2, Table~7 of \citet{gao2026openmopd} and the Qwen
configuration in Appendix~B, Table~3 of \citet{wu2026consolidating},
together with their released implementation defaults.
For every task and backbone, we generate four rollout responses per
training prompt and train for three epochs. All other shared training
hyperparameters remain the same as in the corresponding GRPO
configuration. Table~\ref{tab:opd-training-config} lists the OPD settings
used in this study. The teacher-feedback objective is defined in
Appendix~\ref{app:post-training-objectives}.

\begin{table}[!htbp]
    \centering
    \normalsize
    \setlength{\tabcolsep}{4pt}
    \renewcommand{\arraystretch}{1.08}
    \caption{OPD training configurations.
    Rollout $n$ is the number of responses generated per training prompt.}
    \label{tab:opd-training-config}
    \begin{tabularx}{\linewidth}{@{}l*{4}{>{\centering\arraybackslash}X}@{}}
        \toprule
        & \multicolumn{3}{c}{SmolLM3-3B} & Qwen3-4B \\
        \cmidrule(lr){2-4}
        Hyperparameter & Math & Code & IF & All five domains \\
        \midrule
        Optimizer & AdamW & AdamW & AdamW & AdamW \\
        Learning rate & $10^{-6}$ & $10^{-6}$ & $10^{-6}$ & $10^{-6}$ \\
        Training batch size & 128 & 128 & 128 & 128 \\
        Mini-batch size & 32 & 32 & 32 & 128 \\
        Rollout $n$ & 4 & 4 & 4 & 4 \\
        Training epochs & 3 & 3 & 3 & 3 \\
        Maximum prompt tokens & 1,024 & 2,048 & 2,048 & 5,120 \\
        Maximum response tokens & 30,000 & 30,000 & 2,048 & 16,384 \\
        Rollout temperature & 1.0 & 1.0 & 1.0 & 1.0 \\
        Learning-rate schedule & Constant & Constant & Constant & Constant \\
        Warmup steps & 10 & 10 & 10 & 0 \\
        Gradient clipping & 1.0 & 1.0 & 1.0 & 1.0 \\
        PPO clip (lower / upper) & 0.2 / 0.25 & 0.2 / 0.25 & 0.2 / 0.25 & 0.2 / 0.2 \\
        Reference KL coefficient & 0 & 0 & 0 & 0.001 \\
        Entropy coefficient & 0 & 0 & 0 & 0 \\
        \bottomrule
    \end{tabularx}
\end{table}

The shared implementation settings follow
\href{https://github.com/BytedTsinghua-SIA/Open-MOPD/tree/4809a96cf85a869106ff0ff3f37d0a51e12010ae}{Open-MOPD commit \texttt{4809a96}}
for SmolLM3 and
\href{https://github.com/Di-viner/LLM-Fusion/tree/e314dd28cea056e617490065fc08c4aed90204f3}{LLM-Fusion commit \texttt{e314dd2}}
for Qwen.
We use AdamW with $(\beta_1,\beta_2)=(0.9,0.999)$ and weight decay
$0.01$. For Qwen, the $0.001$ coefficient applies to a separate
reference-policy KL loss.

\subsection{Model Merging}
\label{app:model-merging}

\paragraph{Model Merging.}
We evaluate three representative families of merging operators:
Task Arithmetic (TA), which directly combines task vectors through
scaled summation \citep{ilharco2023taskarithmetic};
TIES-Merging, which introduces coordinate sparsification and sign
conflict resolution \citep{yadav2023ties}; and
TSV-M, which uses an SVD-based transformation before aggregation
\citep{gargiulo2025tsv}.
We also report Raw Sum as the $\lambda=1$ instance of Task
Arithmetic where appropriate.
These methods span increasingly structured approaches to resolving
interactions among task vectors, allowing us to test whether the
observed OPD effects depend on a particular merging rule.

All merges combine parameter-aligned task vectors from the same model
family and shared anchor.
For multi-task experiments, each reported row corresponds to a single
materialized merged checkpoint evaluated across all included domains;
we never select different checkpoints for different benchmarks.

\paragraph{Merging hyperparameters and selection.}
For each merging operator, the reported configuration is selected
as the best-performing one among the evaluated candidates.
Raw Sum uses a global scale of $\lambda=1$.
For the cross-task TSV-M results in
Table~\ref{tab:opd-cross-task-results}, the final merged update is scaled
by $\alpha=0.5$ for SmolLM3 and $\alpha=1.0$ for Qwen.

\section{Complete Parameter-Structure Diagnostics}
\label{app:smollm-structure}
\label{app:geometry-details}

\subsection{Parameter-Structure Measurements}
\label{app:parameter-measurements}

\begin{figure}[!ht]
    \centering
    \includegraphics[width=0.327\linewidth]{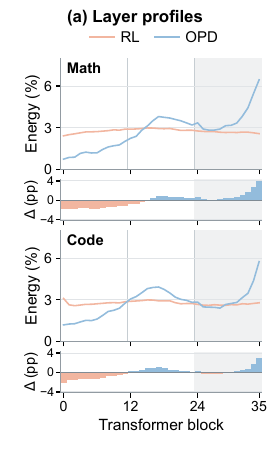}\hfill
    \includegraphics[width=0.327\linewidth]{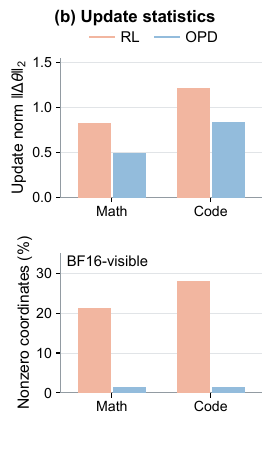}\hfill
    \includegraphics[width=0.327\linewidth]{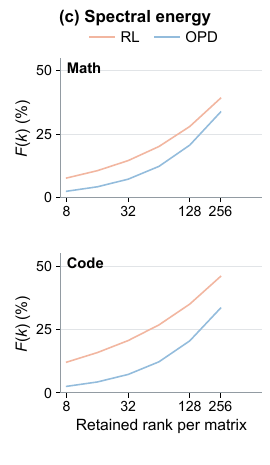}
    \caption{\textbf{Complete parameter-structure profiles for Math and Code.}
    \textbf{(a)} Layer-wise squared $L_2$ norms normalized by full-vector energy,
    with the final 12 Transformer layers shaded; lower strips show OPD minus RL
    in percentage points.
    \textbf{(b)} Task-vector $L_2$ norms (top) and nonzero-coordinate fractions
    in task vectors computed from saved BF16 weights (bottom).
    \textbf{(c)} Spectral concentration $F(k)$, estimated as the fraction of
    total matrix-update energy captured by rank-$k$ approximations across 252 attention
    and MLP projection matrices. The six measured ranks are 8, 16, 32, 64, 128,
    and 256, shown on a logarithmic axis.
    These are descriptive statistics of the evaluated checkpoints.}
    \label{fig:smollm-structure-full}
\end{figure}

For task $t$ and method $m\in\{\mathrm{RL},\mathrm{OPD}\}$, we use
$\Delta\theta_{t,m}=\theta_{t,m}-\theta_0$.
Layer-wise energy is the squared $L_2$ norm normalized by full-vector
energy. The final 12 layers contain $32.1\%\to37.7\%$ of update energy
for Code and $32.3\%\to44.7\%$ for Math, comparing RL with OPD;
the share in the first 12 layers decreases in both tasks.
The OPD $L_2$ norm is $59.7\%$ of RL for Math and $68.7\%$ for Code,
so absolute late-layer energy is lower despite its larger relative share.
BF16-visible nonzero-coordinate fractions are $1.27\%$ versus $21.04\%$
for Math and $1.31\%$ versus $27.98\%$ for Code (OPD versus RL).
These fractions count saved BF16 coordinates that differ from the base,
rather than claiming exact sparsity at higher precision.

We estimate spectral concentration $F(k)$ using approximate randomized
SVD: rank-$k$ retained energies are summed over 252 attention and MLP
projection matrices and divided by their total update energy.
OPD has lower $F(k)$ at all six measured ranks, 8 through 256.
At $k=256$, OPD captures $33.4\%$ of Code energy and $33.7\%$ of Math
energy, versus $45.9\%$ and $39.2\%$ for RL, respectively.
Figure~\ref{fig:smollm-structure-full} provides the complete profiles.

\subsection{Within-Task Diagnostics and Controls}
\label{app:within-task-diagnostics}

\paragraph{Scope of the within-task diagnostics.}
Figure~\ref{fig:within-task-local-alignment}, the orthogonal energy
shares in Section~\ref{sec:within-task-complementarity}, and the Code
controls in Table~\ref{tab:within-task-controls} all use SmolLM3-3B.
Within each task, the diagnostics use the same RL--OPD checkpoint pair;
the Code controls reuse the checkpoints and shared anchor used for the
Code geometry measurements. Table~\ref{tab:within-task-controls} reports
mean score differences relative to the corresponding SmolLM3-3B Raw Sum
model.

\Needspace{7\baselineskip}
\paragraph{A local interpretation of the orthogonal residual.}
For $o=\alpha r+q$ and $\phi=\theta_0+(1+\alpha)r$, a smooth task loss
has the local approximation
$L(\phi+q)-L(\phi)\approx\nabla L(\phi)^\top q+
\frac{1}{2}q^\top H(\phi)q$, where $H$ is the loss Hessian.
Orthogonality to $r$ therefore does not determine the residual's
functional effect; this depends on its interaction with the task loss.
This expression is a local interpretation, not a measured curvature result.
The Code norm-matched controls support a directional contribution,
whereas removing $q$ also changes magnitude and cannot isolate direction.

\paragraph{Qwen within-task performance references.}
Separately, the Qwen3-4B RL specialist means in
Table~\ref{tab:opd-cross-task-results} and within-task Raw Sum means in
Table~\ref{tab:opd-within-task} yield
$59.78-60.14=-0.36$ pp for Math and
$41.01-37.51=+3.50$ pp for Code. These are within-task
comparisons and show that gains vary by task; the cross-task Raw Sum rows of
Table~\ref{tab:opd-cross-task-results} are different compositions.
These Qwen performance comparisons are separate from the SmolLM3-3B
controls in Table~\ref{tab:within-task-controls}.

\subsection{Scale and Localization within Task Hotspots}
\label{app:multitask-diagnostics}

The relative incoming energy in Eq.~\ref{eq:relative-incoming-energy}
factors as $R_t=A_tL_t$. Here $A_t$ measures relative update scale,
and $L_t$ compares the off-task and own-task energy fractions entering
the target hotspots $S_t$. The factorization concerns the sum of
individual incoming energies. Science's ratio decreases
from $3.613$ to $3.262$: its localization factor $L_t$ decreases despite
a slight increase in relative scale $A_t$. Agent's ratio decreases from
$4.637$ to $3.103$: $A_t$ decreases substantially despite higher $L_t$.
Code has reductions in both factors, while the increase in $A_t$
outweighs lower $L_t$ for Math and IF. Jaccard measures shared membership
among selected channels, while $L_t$ measures energy localization.
Direction retention in Eq.~\ref{eq:hotspot-direction-retention} further
captures the alignment among incoming vectors and with the target-task
update through their actual vector sum.

\section{Additional Context on OPD and Capability Integration}
\label{app:opd-related-work}

\paragraph{On-Policy Distillation and Capability Integration.}
On-policy distillation (OPD) trains a student using teacher feedback on
student-generated trajectories, reducing the mismatch between training
sequences and inference-time behavior. Generalized Knowledge Distillation
(GKD; \citealp{agarwal2024gkd}) studies this approach with flexible
teacher--student divergence objectives, while MiniLLM~\citep{gu2024minillm}
develops reverse-KL optimization for generative language-model distillation.
Beyond single-teacher transfer, MOPD~\citep{ma2026mopd} consolidates
domain-specific RL teachers into a shared student through token-level
supervision on its own rollouts. Open-MOPD~\citep{gao2026openmopd} examines
capability imbalance in multi-teacher distillation, identifying token-level
optimization-budget misallocation as a major bottleneck in its controlled
setting and improving integration through budget balancing, adaptive
allocation, and reward refresh. Complementing these training-time
integration approaches, we examine the parameter updates produced by
task-specific OPD and their reuse as task vectors for subsequent composition
in parameter space.

\end{document}